\documentclass[sigconf,screen=true,anonymous=false,9pt]{acmart}
\usepackage{varwidth}
\usepackage{tcolorbox}

\renewcommand\footnotetextcopyrightpermission[1]{} 
\usepackage{lipsum}
\usepackage{titlesec}

\usepackage{graphicx}
\usepackage{amsmath}
\usepackage{mathtools}
\usepackage{comment}
\usepackage[subrefformat=parens,labelformat=parens]{subfig}
\usepackage{bm}
\usepackage{multirow}
\usepackage{threeparttable,booktabs}
\usepackage{blkarray}
\usepackage{tikz}
\usepackage{balance}
\usepackage{courier}                                       
\usepackage{cleveref}                                      
\usepackage[mathcal]{eucal}
\usepackage[noend]{algpseudocode}
\algrenewcommand\textproc{\texttt}
\makeatletter\let\float@addtolists\relax\makeatother
\usepackage{algorithm}

\usepackage{filecontents}                                  
\usepackage{pgfplots}
\usepackage{pgfplotstable}
\pgfplotsset{compat=newest}
\usepackage[figuresright]{rotating}
\usetikzlibrary{positioning}
\usetikzlibrary{fit}

\renewcommand{\vec}[1]{\boldsymbol{#1}}

\theoremstyle{plain}

\theoremstyle{definition}
\newtheorem{mydefinition}{\textbf{Definition}}

\graphicspath{{./figs/}}

\definecolor{NVgreen}{RGB}{118,185,0}
\definecolor{NVblack}{RGB}{0,0,0}
\definecolor{NVlgrey}{RGB}{205,205,205}
\definecolor{NVmgrey}{RGB}{140,140,140}
\definecolor{NVdgrey}{RGB}{94,94,94}

\definecolor{NVemerald}{RGB}{0,133,100}
\definecolor{NVamethyst}{RGB}{93,22,130}
\definecolor{NVintel}{RGB}{0,113,197}
\definecolor{NVgarnet}{RGB}{137,12,88}
\definecolor{NVfluorite}{RGB}{250,194,0}

\newcommand\lithobench{\texttt{LithoBench}}

\begin{document}

\title{
   Pushing the Limits of Inverse Lithography with Generative Reinforcement Learning
}

\author{Haoyu Yang}
\affiliation{%
  \institution{NVIDIA Corp.}
  \city{Austin}
  \state{TX}
  \country{USA}
}
\email{haoyuy@nvidia.com}

\author{Haoxing Ren}
\affiliation{%
  \institution{NVIDIA Corp.}
  \city{Austin}
  \state{TX}
  \country{USA}
}
\email{haoxingr@nvidia.com}

\begin{abstract}
Inverse lithography (ILT) is critical for modern semiconductor manufacturing but suffers from highly non-convex objectives that often trap optimization in poor local minima. 
Generative AI has been explored to warm-start ILT, yet most approaches train deterministic image-to-image translators to mimic sub-optimal datasets, providing limited guidance for escaping non-convex traps during refinement.
We reformulate mask synthesis as conditional sampling: a generator learns a distribution over masks conditioned on the design and proposes multiple candidates.
The generator is first pretrained with WGAN plus a reconstruction loss, then fine-tuned using Group Relative Policy Optimization (GRPO) with an ILT-guided imitation loss.
At inference, we sample a small batch of masks, run fast batched ILT refinement, evaluate lithography metrics (e.g., EPE, process window), and select the best candidate.
On \texttt{LithoBench} dataset, the proposed hybrid framework reduces EPE violations under a 3\,nm tolerance and roughly doubles throughput versus a strong numerical ILT baseline, while improving final mask quality.
We also present over 20\% EPE improvement on \texttt{ICCAD13} contest cases with 3$\times$ speedup over the SOTA numerical ILT solver.
By learning to propose ILT-friendly initializations, our approach mitigates non-convexity and advances beyond what traditional solvers or GenAI can achieve.
\end{abstract}
\maketitle
\renewcommand{\shortauthors}{Haoyu Yang \& Haoxing Ren}

\section{Introduction}
\label{sec:intro}
Inverse lithography technology (ILT) is a key enabler for advanced semiconductor manufacturing, 
thanks to its ability to deliver larger process windows and GPU-friendly optimization efficiency.
However, ILT remains challenging due to highly non-convex objectives that often trap optimization in poor local minima.
To address this, two main research directions have emerged: (1) improving numerical ILT solvers \cite{OPC-DAC2014-Gao,OPC-DAC2022-Wang,OPC-DAC2023-Sun,curvyilt,OPC-DATE2021-Yu} and (2) leveraging generative AI to warm-start ILT \cite{OPC-ICCAD2020-DAMO,ililt,OPC-TCAD2020-Yang}.

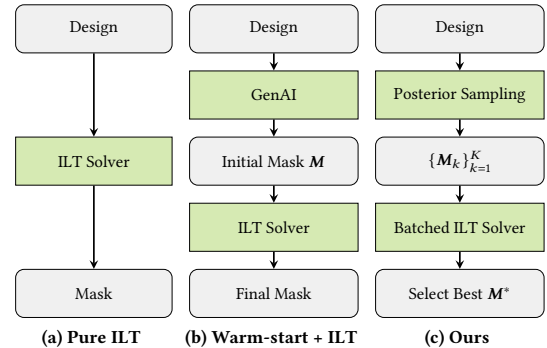
\begin{figure}[t]
	\centering
	\usetikzlibrary{arrows.meta,positioning,fit}
	\providecommand{\stylearchscale}{0.8}
	
	\subfloat[Pure ILT]{
		\scalebox{\stylearchscale}{
		\begin{tikzpicture}[node distance=2.2cm, font=\small, baseline=(current bounding box.north)]
			\tikzstyle{startstop} = [rectangle, rounded corners, minimum width=2.6cm, minimum height=0.8cm, text centered, draw=black, fill=NVlgrey!30]
			\tikzstyle{process} = [rectangle, minimum width=2.6cm, minimum height=0.8cm, text centered, draw=black, fill=NVgreen!30]
			\tikzstyle{arrow} = [thick,->,>=stealth]
			
			\node (design) [startstop] {Design};
			\node (solver) [process, below of=design] {ILT Solver};
			\node (mask) [startstop, below of=solver] {Mask};
			
			\draw [arrow] (design) -- (solver);
			\draw [arrow] (solver) -- (mask);
		\end{tikzpicture}}
	}
	\subfloat[Warm-start + ILT]{
		\scalebox{\stylearchscale}{
		\begin{tikzpicture}[node distance=1.1cm, font=\small, baseline=(current bounding box.north)]
			\tikzstyle{startstop} = [rectangle, rounded corners, minimum width=2.8cm, minimum height=0.8cm, text centered, draw=black, fill=NVlgrey!30]
			\tikzstyle{process} = [rectangle, minimum width=2.8cm, minimum height=0.8cm, text centered, draw=black, fill=NVgreen!30]
			\tikzstyle{arrow} = [thick,->,>=stealth]
			
			\node (design2) [startstop] {Design};
			\node (genai) [process, below of=design2] {GenAI};
			\node (initmask) [startstop, below of=genai] {Initial Mask $\vec{M}$};
			\node (solver2) [process, below of=initmask] {ILT Solver};
			\node (finalmask) [startstop, below of=solver2] {Final Mask};
			
			\draw [arrow] (design2) -- (genai);
			\draw [arrow] (genai) -- (initmask);
			\draw [arrow] (initmask) -- (solver2);
			\draw [arrow] (solver2) -- (finalmask);
		\end{tikzpicture}}
	}
	\subfloat[Ours]{
		\scalebox{\stylearchscale}{
		\begin{tikzpicture}[node distance=1.1cm, font=\small, baseline=(current bounding box.north)]
			\tikzstyle{startstop} = [rectangle, rounded corners, minimum width=2.8cm, minimum height=0.8cm, text centered, draw=black, fill=NVlgrey!30]
			\tikzstyle{process} = [rectangle, minimum width=2.8cm, minimum height=0.8cm, text centered, draw=black, fill=NVgreen!30]
			\tikzstyle{arrow} = [thick,->,>=stealth]
			
			\node (design3) [startstop] {Design};
			\node (sampler) [process, below of=design3] {Posterior Sampling};
			
			\node (mcands) [startstop, below of=sampler] {$\{\vec{M}_k\}_{k=1}^{K}$};
			
			\node (ilt) [process, below of=mcands] {Batched ILT Solver};
			\node (select) [startstop, below of=ilt] {Select Best $\vec{M}^*$};
			
			\draw [arrow] (design3) -- (sampler);
			\draw [arrow] (sampler) -- (mcands);
			
			\draw [arrow] (mcands) -- (ilt);
			\draw [arrow] (ilt) -- (select);
		\end{tikzpicture}}
	}
	
	\caption{Comparison of ILT schemes: (a) Pure ILT with iterative updates; (b) GenAI warm-start followed by ILT refinement; (c) Proposed conditional mask sampling that generates multiple candidates, applies batched fast ILT refinement, and selects the best result.}
	\label{fig:intro-ilt-schemes}
\end{figure}

ILT solvers have been extensively studied over the past decades. MOSAIC \cite{OPC-DAC2014-Gao} was one of the earliest academic attempts to develop a gradient-based ILT with explicit EPE-violation considerations. 
However, the CPU-centric implementation limited efficiency. 
NeuralILT \cite{OPC-ICCAD2020-NeuralILT} accelerates the flow by moving lithography operators to GPUs, yielding significant speedups and incorporating mask-rule constraints into optimization.
\cite{OPC-DATE2021-Yu} introduces mask smoothness by formulating mask representations with level-set functions, significantly reducing e-beam shots. 
A2ILT \cite{OPC-DAC2022-Wang} incorporates a spatial attention map to constrain ILT gradients and avoid generating mask-rule-violation artifacts.
MultiILT \cite{OPC-DAC2023-Sun} introduces an offset to the design image, which stimulates the generation of sub-resolution assist features (SRAFs) and leads the leaderboard on the \texttt{ICCAD13} benchmark suite. 
Recently, as the multi-beam mask writer (MBMW) enters production \cite{mbmw}, curvilinear masks have regained research attention.
CurvyILT \cite{curvyilt} is the first ILT implementation targeting curvilinear mask generation with a differentiable morphological operator, directly producing MRC-clean results while achieving state-of-the-art performance on open lithography benchmarks \cite{lithobench,OPC-ICCAD2013-Banerjee}.
However, most solver-focused efforts overlook the strong dependence on initialization arising from ILT's non-convexity. 

Generative AI has therefore been explored to warm-start ILT, aiming to provide better initializations that reduce optimization iterations while improving final mask QoR. 
GAN-OPC \cite{OPC-TCAD2020-Yang,OPC-DAC2018-Yang} was the first to introduce generative AI to produce initial masks, significantly reducing MOSAIC iterations with better QoR. 
Subsequent works mainly focus on architectural improvements, including DAMO \cite{OPC-ICCAD2020-DAMO}, which leverages Pix2Pix \cite{pix2pix} for design-to-mask translation.
\cite{CFNO} develops physics-inspired modeling with a customized Fourier Neural Operator that produces masks competitive with or better than the numerical solver A2ILT.
ILILT \cite{ililt} reformulates the ILT problem as implicit-layer learning, further boosting mask generation but incurring heavier inference cost.
However, most of these approaches train deterministic image-to-image translators to mimic sub-optimal datasets, providing limited guidance for escaping non-convex traps during refinement.

We argue that the traditional scheme of \textbf{AI warm-up + ILT refinement} is sub-optimal; ILT should be formulated as a conditional sampling task given the non-convex nature of mask optimization. 
Specifically, \textbf{a generative model should learn a distribution over masks conditioned on the design and propose multiple candidates,
which then serve as seeds for ILT refinement} (see \Cref{fig:intro-ilt-schemes}).
However, this formulation faces two challenges: 
1) the generator must serve as an effective posterior sampler that yields a distribution of reasonable masks rather than adversarial artifacts;
2) the samples should consistently benefit subsequent ILT refinement steps.

In this paper, we tackle these challenges with a dedicated generative architecture and a targeted training flow. 
First, we propose a style-aware generation plugin that can be integrated into neural network models to control sampling style while maintaining the true design manifold.
To ensure effective posterior sampling, we adopt a two-stage training mechanism comprising generative pretraining and reinforcement fine-tuning. 
Specifically, the pretraining stage teaches the model the mask distribution, and the second stage fine-tunes the generator to a conditional distribution such that sampled masks have a higher likelihood of reaching high-quality optima after ILT refinement. 
This is achieved with a customized reinforcement learning algorithm leveraging Group Relative Policy Optimization (GRPO) and imitation learning. 
Major contributions of the paper include:
\begin{itemize}
    \item We revisit GenAI-aided ILT by treating the generator as a posterior sampler explicitly trained for ILT refinement. 
    \item We propose a style-aware plugin for neural network architectures that can produce variations of masks conditioned on post-PnR layouts.
    \item We propose a two-stage training mechanism with pretraining and reinforcement learning to ensure the generative model captures the true mask design manifold while achieving effective posterior sampling for ILT refinement. 
    \item We conduct extensive experiments on open benchmarks from \texttt{LithoBench} and the \texttt{ICCAD13} CAD contest, and for the first time our method reduces EPE violations below 3\,nm on a large fraction of cases with 3$\times$ speedup.
\end{itemize}

The remainder of the paper is organized as follows: 
\Cref{sec:prelim} covers terminology related to inverse lithography and related work;
\Cref{sec:alg} details the methodology and implementation of the proposed framework;
\Cref{sec:result} presents experimental results to support our methodology and claims,
followed by the conclusion and future work in \Cref{sec:conclu}.

\section{Preliminaries}
\label{sec:prelim}

\begin{figure}
    \centering
    \providecommand{\stylearchscale}{0.8}
	\scalebox{\stylearchscale}{
    \begin{tikzpicture}[scale=1.0]
    
    
    \draw[line width=0.8pt] (0,0) rectangle (4.6,1.4);
    
    \draw[NVgreen, line width=1pt]
        (2.3,0.7) ellipse [x radius=2.2, y radius=0.55];
    
    \draw[red, line width=1.2pt] (3.55,0) -- (3.55,0.25);

    
    \begin{scope}[yshift=-1.8cm]
    
    \draw[line width=0.8pt] (0,0) rectangle (4.6,1.4);
    
    \path (2.3,0.7) coordinate (C);
    
    \draw[NVlgrey, line width=1pt]
        (C) ellipse [x radius=2.2, y radius=0.55];
    
    \draw[NVlgrey, line width=1pt]
        (C) ellipse [x radius=1.8, y radius=0.35];
    
    \begin{scope}
        \clip (C) ellipse [x radius=2.2, y radius=0.55];
        \fill[NVlgrey] (C) ellipse [x radius=2.2, y radius=0.55];
        \fill[white] (C) ellipse [x radius=1.8, y radius=0.35];
    \end{scope}
    
    \end{scope}

    
    \begin{scope}[xshift=5.cm, yshift=-.4cm]
    
    \draw[line width=0.8pt] (0,0) rectangle ++(0.45,0.2);
    \node[right] at (0.6,0.1) {Design};
    
    \draw[NVgreen, line width=1pt] (0.22,-0.3) ellipse [x radius=0.24, y radius=0.13];
    \node[right] at (0.6,-0.3) {Contour};
    
    \draw[red, line width=1pt] (0,-0.7) -- ++(0.5,0);
    \node[right] at (0.6,-0.7) {EPEV};
    
    \fill[NVlgrey, line width=1pt] (0.22,-1.1) ellipse [x radius=0.24, y radius=0.13];
    \node[right] at (0.6,-1.1) {PVB};
    
    \end{scope}
    
    \end{tikzpicture}}
    
    \caption{Mask optimization metrics. Resketched from \cite{curvyilt}.}
    \label{fig:metric}
\end{figure}
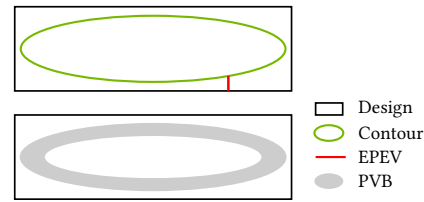

Lithography is the critical step in chip manufacturing, where physical-implemented designs will be transferred onto silicon wafers. 
However, due to the mismatch between the design features and the capabilities of the lithography system (source wavelength, numerical aperture, etc),
the photo mask $\vec{M}$ of the design $\vec{Z}^\ast$ should be optimized to compensate distortion. 
In practical, photolithography is super complicated that requires regiorous EM simulation, for proof of concept and follow the practice of ILT in academic literature, we adopt the abstracted Sum of Coherent Systems (SOCS) model in this paper. 
Fundamentally, the forward imaging models how is a mask being printed on the silicon, as in \Cref{eq:socs} \cite{DFM-B2011-Ma}.
\begin{align}
    \label{eq:socs}
    \vec{I} = \sum_{1}^{i=k} \alpha_i ||\vec{M}\otimes \vec{h}_i||^2,
\end{align}
where $\alpha_i$'s and $\vec{h}_i$'s are eigen values and eigen vectors of the transmission coefficent matrixs, respectively,
and $\vec{I}$ is the aerial image projected on the resist materials. 
There would also be models related to resist behavior with constant thresholding. 
\begin{align}
    \label{eq:resist}
    \vec{Z}_{i,j} = 
    \begin{cases}
        1 & \text{if}~~\vec{I}_{i,j} > I_\text{th}, \\
        0 & \text{otherwise},\\
    \end{cases}
\end{align}
where $\vec{Z}$ denots the resist contour on the silicon wafer. 
\textit{ILT is to use numerical methods to find the mask $\vec{M}$ such that the remaining pattern on the silicon wafer $\vec{Z}$ after lithography is as close as the original design $\vec{Z}^\ast$ at various manufacturing conditions.}
We use the academic accepted metrics \textit{Edge Placement Error (EPE)} and \textit{Process Viaration Band (PVB) Area} to reflect this objective, as shown in \Cref{fig:metric}.
It should be noted that we will use curvilinear mask assumption and its associated MRC rules defined in \cite{curvyMRC} and we will enforce MRC-cleaness by morphological operations as suggested in \cite{curvyilt}.

\begin{mydefinition}[EPE \cite{curvyilt,OPC-ICCAD2013-Banerjee}]
Edge placement error (EPE) quantifies the distance between the edge of the target design and the edge of the actual printed feature on the wafer. When this distance exceeds a certain threshold, typically a few nanometers, the design is at risk of failing. Each instance where this threshold is surpassed is referred to as an EPE violation. A well-optimized mask should minimize the occurrence of these EPE violations as much as possible.
\end{mydefinition}
\begin{mydefinition}[PVB \cite{curvyilt,OPC-ICCAD2013-Banerjee}]
The process variation band (PVB) illustrates how the printed wafer image fluctuates due to variations
in the manufacturing process. A common approach to quantitatively assess this variation involves perturbing simulation
parameters related to system settings, such as the lens focus plane and UV dose strength. The area between the innermost
and outermost contours of the printed image represents the PVB. A smaller PVB indicates greater robustness of the mask
against process variations.
\end{mydefinition}

\section{The Framework}
\label{sec:alg}

\subsection{Revisiting Generative AI for ILT}
In recent years, vision-based generative models have been extensively studied for computational lithography problems, including aerial image prediction 
\cite{DFM-DAC2019-Ye,DFM-DAC2022-Yang,OPC-ICCAD2020-DAMO} and inverse lithography \cite{OPC-TCAD2020-Yang,OPC-ICCAD2020-DAMO,OPC-ICCAD2020-NeuralILT,ililt}.  
For inverse modeling in particular, a common training practice is to solve a minimization task that reduces the gap between the model-generated mask and the ground-truth training mask.  
Although advances in model architectures and loss functions have improved performance and reduced the number of post-ILT refinement iterations, two critical issues remain unresolved, leaving generative AI–based inverse solutions sub-optimal:
1) The training dataset is often sub-optimal as examplified in \texttt{LithoBench} \cite{lithobench}; simply mimicking its behavior does not necessarily yield higher-quality masks. 
2) Post-training ILT refinement involves navigating a highly non-convex manifold. Without explicit guidance, the generative model struggles to achieve fast convergence or escape poor local optima, which limits current state-of-the-art methods to performance levels still bounded by pure numerical ILT solutions.

\subsection{Generative AI as Mask Sampler}
We argue that the limitations of prior generative approaches to ILT stem from relying on deterministic image-to-image translation.  
Instead, we model a learned distribution over the mask space conditioned on designs, i.e., $\vec{M} \sim P_{\vec{M} \mid \vec{Z}}$,  
where $\vec{Z}$ denotes the Manhattan design vector and $\vec{M}$ denotes the mask vector.  
To approximate this conditional distribution, we train a conditional generator $G$ that takes as input a design $\vec{Z} \sim P_{\vec{Z}}$ and a random noise vector $\vec{q} \sim \mathcal{N}$.  
Formally, the generated mask is given by
\begin{align}
	\vec{M} \sim P_{\vec{M} \mid \vec{Z}} \approx (G(\vec{Z}, \cdot))_{\#} \mathcal{N}, 
\end{align}
where $(G(\vec{Z}, \cdot))_{\#} \mathcal{N}$ denotes the pushforward of the noise distribution $\mathcal{N}$ through the generator $G$ conditioned on $\vec{Z}$.  
Note that the learned distribution $P_{\vec{M} \mid \vec{Z}}$ does not necessarily align with the optimal mask manifold.  
With this formulation, the generator does not produce a single mask but a distribution of masks for any given design, serving as a sampler whose proposals are subsequently refined by ILT.  
Given a design $\vec{Z}$, $G$ should sample a small set of plausible masks that are amenable to rapid refinement toward high-quality solutions.
Before detailing how we train $G$ to achieve this, we first describe the architecture.

\subsubsection{Style-Aware Generative Model Architecture}
\begin{figure*}[t]
	\centering
	\usetikzlibrary{arrows.meta,positioning,fit,shapes.misc}

        \providecommand{\stylearchscale}{0.8}
	\scalebox{\stylearchscale}{
	\begin{tikzpicture}[node distance=0.35cm, font=\small]
		
		\tikzstyle{data}    = [rectangle, rounded corners, minimum width=2.2cm, minimum height=0.7cm, text centered, align=center, text width=2.3cm, draw=black, fill=NVlgrey!30]
		\tikzstyle{module}  = [rectangle, minimum width=2.4cm, minimum height=0.8cm, text centered, align=center, text width=2.5cm, draw=black, fill=NVgreen!30]
		\tikzstyle{stylemod}= [rectangle, minimum width=2.4cm, minimum height=0.8cm, text centered, align=center, text width=2.5cm, draw=black, fill=NVamethyst!20]
		\tikzstyle{adain}   = [rectangle, minimum width=2.2cm, minimum height=0.7cm, text centered, align=center, text width=2.3cm, draw=black, fill=NVintel!20]
		\tikzstyle{arrow}   = [thick,->,>=stealth,shorten >=1pt,shorten <=1pt]
		\tikzstyle{group}   = [dashed, draw=NVmgrey, rounded corners, inner sep=0.25cm]
		\tikzstyle{op}      = [circle, minimum size=0.6cm, draw=black, fill=NVlgrey!30]
		
		\node (Zfull) [data] {Design $\vec{Z}$\\(full)};
		\node (head)  [module, below=0.5cm of Zfull] {Head Downsample\\(optional)};
		\node (Z0)    [data, below=0.35cm of head] {$\vec{Z}_{0}$};
		\node (down1) [module, below=0.35cm of Z0] {Downsample};
		\node (Z1)    [data, below=0.35cm of down1] {$\vec{Z}_{1}$};
		\node (down2) [module, below=0.35cm of Z1] {Downsample};
		\node (Z2)    [data, below=0.35cm of down2] {$\vec{Z}_{2}$};
		\node (pyrbox) [group, fit=(Zfull) (head) (Z0) (down1) (Z1) (down2) (Z2)] {};
		\node[above=0.05cm of pyrbox, text=NVdgrey] {Input Pyramid};
		\draw [arrow] (Zfull) -- (head);
		\draw [arrow] (head) -- (Z0);
		\draw [arrow] (Z0) -- (down1);
		\draw [arrow] (down1) -- (Z1);
		\draw [arrow] (Z1) -- (down2);
		\draw [arrow] (down2) -- (Z2);
		
		\node (encd)   [module, right=1cm of Z2] {Downsample};
		\node (bott)   [stylemod, right=0.35cm of encd]  {Style ResBlocks $\times n$};
		\node (decu)   [module, right=0.35cm of bott] {UpConv};
		\node (Y2)     [data, right=0.35cm of decu] {Low Mask};
		\node (globbox) [group, fit=(encd) (bott) (decu) (Y2)] {};
		\node[above=0.05cm of globbox, text=NVdgrey] {Level 2};
		
		
		\draw [arrow] (Z2) -- (encd);
		\draw [arrow] (encd) -- (bott);
		\draw [arrow] (bott) -- (decu);
		\draw [arrow] (decu) -- (Y2);
		
		\node (z)   [data, below=0.6cm of Z2] {Latent\\$\vec{z}\!\sim\!\mathcal{N}(0,\mathbf{I})$};
		\node (map) [stylemod, right=1cm of z] {Style Mapping\\$f_{\phi}$};
		\node (w)   [data, below=0.6cm of bott] {Style\\$\vec{w}$};
		\draw [arrow] (z) -- (map);
		\draw [arrow] (map) -- (w);
        \draw [arrow] (w) -- (bott);
		
		\node (LE1d) [module, right=1.0cm of Z1] {Local\\Downsample};
		\node (add1) [op, right=0.35cm of LE1d] {$+$};
		\node (LE1r) [module, right=0.35cm of add1] {Local ResBlocks};
		\node (LE1u) [module, right=0.35cm of LE1r] {UpConv};
		\node (Y1)   [data, right=0.35cm of LE1u] {Low Mask};
		\node (le1box) [group, fit=(LE1d) (add1) (LE1r) (LE1u) (Y1)] {};
		\node[above=0.05cm of le1box, text=NVdgrey] {Level 1};
		\path (Y2.north) ++(0,0.95cm) coordinate (route1top);
		\path (add1.south) ++(0,-0.55cm) coordinate (route1right);
		\draw [arrow] (Z1) -- (LE1d);
		\draw [arrow] (LE1d) -- (add1);
		\draw [arrow] (Y2.north) -- (route1top) -- (route1right) -- (add1.south);
		\draw [arrow] (add1) -- (LE1r);
		\draw [arrow] (LE1r) -- (LE1u);
		\draw [arrow] (LE1u) -- (Y1);
		
		\node (LE2d) [module, right=1.0cm of Z0] {Local\\Downsample};
		\node (add2) [op, right=0.35cm of LE2d] {$+$};
		\node (LE2r) [module, right=0.35cm of add2] {Local ResBlocks};
		\node (LE2u) [module, right=0.35cm of LE2r] {UpConv};
		\node (Y0)   [data, right=0.35cm of LE2u] {$\vec{M} \sim P_{\vec{M}|\vec{Z}}$};
		\node (le2box) [group, fit=(LE2d) (add2) (LE2r) (LE2u) (Y0)] {};
		\node[above=0.05cm of le2box, text=NVdgrey] {Level 0};
		\path (Y1.north) ++(0,0.95cm) coordinate (route2top);
		\path (add2.south) ++(0,-0.55cm) coordinate (route2right);
		\draw [arrow] (Z0) -- (LE2d);
		\draw [arrow] (LE2d) -- (add2);
		\draw [arrow] (Y1.north) -- (route2top) -- (route2right) -- (add2.south);
		\draw [arrow] (add2) -- (LE2r);
		\draw [arrow] (LE2r) -- (LE2u);
		\draw [arrow] (LE2u) -- (Y0);

    \path (Z0.east) ++(16.2cm, 1cm) coordinate (rb_anchor);
    \node (rb_x)   [data, at={(rb_anchor)}] {$\vec{x}$};
    \node (rb_ad1) [adain, below=0.25cm of rb_x] {AdaIN($\vec{w}$)};
    \node (rb_c1)  [module, below=0.22cm of rb_ad1] {Conv};
    \node (rb_ad2) [adain, below=0.22cm of rb_c1] {AdaIN($\vec{w}$)};
    \node (rb_c2)  [module, below=0.22cm of rb_ad2] {Conv};
    \node (rb_plus)[op, below=0.25cm of rb_c2] {$+$};
    \node (rb_y)   [data, below=0.25cm of rb_plus] {$\vec{y}$};
    \draw [arrow] (rb_x) -- (rb_ad1) -- (rb_c1) -- (rb_ad2) -- (rb_c2) -- (rb_plus) -- (rb_y);
    \path (rb_x.east) ++(0.3cm,0) coordinate (rb_skip_right);
    \draw [arrow] (rb_x.east) -- (rb_skip_right) |- (rb_plus.east);
    \node (rb_box) [group, fit=(rb_x) (rb_ad1) (rb_c1) (rb_ad2) (rb_c2) (rb_plus) (rb_y)] {};
    \node[above=0.05cm of rb_box, text=NVamethyst] {Style ResBlock (detail)};
    \end{tikzpicture}
    }
	\caption{Detailed style-aware generator. A pyramid of inputs $\{\vec{Z}_\ell\}$ is formed (optional head downsample). The coarsest input drives the level 2 generator: stride-2 downsamples, style-modulated residual trunk (AdaIN with $\vec{w}=f_\phi(\vec{z})$), then symmetric upsampling and bounded output. Two other levels operate coarse-to-fine by fusing the upsampled output from the previous stage with downsampled features at the current resolution (element-wise addition), refining via local ResBlocks and upsampling. An optional final bicubic interpolation restores the original resolution.}
	\label{fig:style-arch}
\end{figure*}
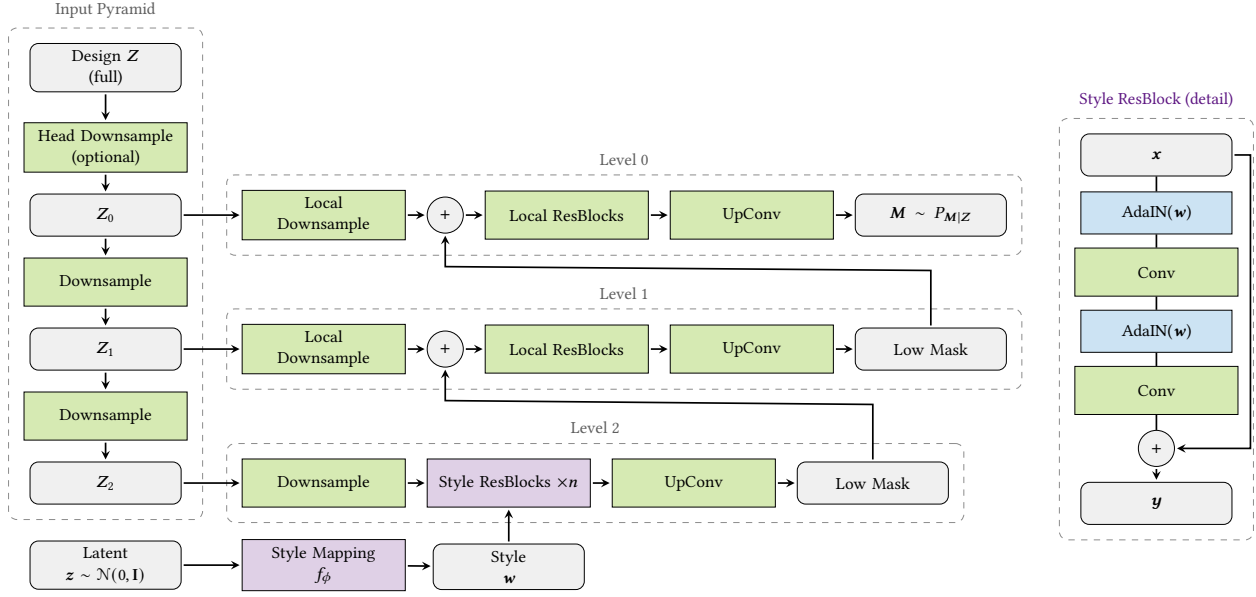

Two common posterior modeling families are: (1) GANs that map noise to in-distribution samples, and (2) VAEs whose outputs parameterize distributions $\mathcal{N}(\vec{\mu}, \vec{\sigma}^2)$.  
Neither directly satisfies our objectives. Vanilla conditional GANs can suffer from mode collapse and often need heavy regularization and pristine data to maintain diversity under design conditioning.  
VAEs, while stable, tend to produce blurry or noisy outputs at very high resolutions (e.g., 2048$\times$2048).  
Inspired by StyleGAN \cite{stylegan}, we inject noise in a coarse latent space and modulate a content-preserving backbone with style, yielding a style-aware generator (Fig. \ref{fig:style-arch}). The architecture follows a multi-scale path: strided convolutions for downsampling, content encoding via repeated ResNet blocks, and transposed convolutions for upsampling.  
Our core innovation is a customized Style ResBlock at the coarse level (Level 2) that applies adaptive instance normalization (AdaIN) using a style code $\vec{w}$:
\begin{align}
	\label{eq:adain}
	\operatorname{AdaIN}(\vec{x}; \vec{w}) 
	= \gamma(\vec{w}) \odot \dfrac{\vec{x} - \mu(\vec{x})}{\sigma(\vec{x}) + \epsilon} + \beta(\vec{w}),
\end{align}
where a lightweight MLP $f_{\phi}$ maps latent noise $\vec{z} \sim \mathcal{N}(\vec{0}, \mathbf{I})$ to a style code $\vec{w} = f_{\phi}(\vec{z})$;  
$\mu(\cdot)$ and $\sigma(\cdot)$ are per-channel statistics, and $[\gamma(\vec{w}), \beta(\vec{w})]$ are affine transforms of $\vec{w}$.  
This decouples \emph{content} (from $\vec{Z}$) and \emph{style} (from $\vec{w}$): sampling perturbs mask style while preserving design semantics.

\noindent\textit{Why style-space sampling helps?}
Sampling in style space provides a low-dimensional, well-conditioned exploration knob with smooth control over variations.  
Because features are anchored to $\vec{Z}$ and style enters only via AdaIN, generated masks stay design-consistent and avoid off-manifold artifacts common with direct pixel-space sampling.

\subsubsection{Design insights.}
Practically, we encode the design once and let $\vec{w}$ modulate only normalized features in the style blocks; this preserves mask topology and prevents structural drift. During pretraining, WGAN with a reconstruction term aligns the learned manifold with valid masks, while light priors (e.g., smoothness) discourage spurious high-frequency artifacts. For exploration, we treat $\vec{z}$ as a reparameterized Gaussian and control its scale (or truncation in $\vec{w}$) to anneal diversity across training stages. At inference and in RL fine-tuning, we generate a small group of proposals per design, apply fast ILT to each, and compute group-relative advantages (GRPO) to update $G$ stably with low-variance signals.
Overall, the style-aware generator provides controllable, design-faithful sampling that stays on-mask-manifold and integrates naturally with GRPO and imitation for ILT-focused fine-tuning.

\subsection{Training Methodology}
Recall we have set two major objectives of $G$:
(1) $G$ should be an conditional posterior sampler that corresponds to a distribution of plausible masks given any design $\vec{Z}$ and
(2) $G$ should be able to produce masks that are likely to reach optimal after ILT refinement. 
It is not trivial to train $G$ to achieve both. 
Here inspired by the general LLM training paradigm, we propose a two-stage training mechanism: generative pre-training (PT) and reinforcement finetuning,
where pre-training fills the generative model with prior knowledges on the mask design manifold, and finetuning guides the generator to sample masks that are likely to reach optimal after ILT refinement.

\subsubsection{Generative Pre-Training}
During the PT phase, 
the generator $G$ is trained with a combination of the Wasserstein GAN loss 
and a reconstruction loss on the mask space.  
The overall objective is formulated as
\begin{align}
	\label{eq:pt-objective}
	\min_{G} \max_{D} \; 
	&\mathbb{E}_{\vec{M} \sim P_{\vec{M} \mid \vec{Z}}} \big[ D(\vec{M}, \vec{Z}) \big] \\ \nonumber
	- &\mathbb{E}_{\vec{Z} \sim P_{\vec{Z}}, \; \vec{q} \sim \mathcal{N}} 
	\big[ D(G(\vec{Z}, \vec{q}), \vec{Z}) \big] \\ \nonumber
	+ &\lambda_1 \, \mathbb{E}_{\vec{M}, \vec{Z}} 
	\big[ \ell_{\text{rec}}(G(\vec{Z}, \vec{q}), \vec{M}) \big] \\ \nonumber
	- &\lambda_2 \, \mathbb{E}_{\hat{\vec{M}}, \vec{Z}} 
	\big[ \big( \lVert \nabla_{\hat{\vec{M}}} D(\hat{\vec{M}}, \vec{Z}) \rVert_{2} - 1 \big)^{2} \big],
\end{align}
where $D$ is the ordinary discriminator (or critic) in GANs designed with gradient penalty \cite{wgan}, 
$\ell_{\text{rec}}$ is the reconstruction loss (e.g., $\ell_{1}$ or $\ell_{2}$), 
and $\lambda_1 > 0$ and $\lambda_2 > 0$ balance adversarial learning with mask reconstruction fidelity and gradient penalty, respectively. 
Here, $\hat{\vec{M}} = \epsilon \, \vec{M} + (1-\epsilon) \, G(\vec{Z}, \vec{q})$ with $\epsilon \sim \mathcal{U}(0,1)$ is the linearly interpolated mask used for the WGAN-GP term, and $\lambda_2 > 0$ weights the gradient penalty. 
After the PT phase, $G$ is almost at the stage of prior arts. It could also served as a policy network that produces actions (masks).

\subsubsection{Reinforcement Finetuning}
We fine-tune $G$ to sample masks that remain close to the design manifold and are likely to converge to high quality after a short ILT refinement. 
For a given design $\vec{Z}$, we draw a group of latents $\{\vec{q}_{k}\}_{k=1}^{K} \sim \mathcal{N}$ and generate logits $\{\vec{Y}_{k}\}_{k=1}^{K}$, where $\vec{Y}_{k} = G(\vec{Z}, \vec{q}_{k})$. 
We form binary masks as actions by thresholding, $\vec{M}_{k} = \mathbb{I}[\vec{Y}_{k} > 0.5]$. 
Each $\vec{M}_{k}$ is passed through a few-step, low-resolution ILT loop and then upsampled to full resolution to obtain $\vec{M}^{\text{ILT}}_{k}$; 
the reward is the negative edge placement error (EPE) after refinement, $R_{k} = -\operatorname{EPE}(\vec{M}^{\text{ILT}}_{k}, \vec{Z})$. 

\paragraph{Teacher-relative group policy optimization.}
Group Relative Policy Optimization (GRPO) has recently gained popularity in LLM finetuning tasks.
Original GRPO computes a group-relative baseline $R_{k} - \frac{1}{K}\sum_{j=1}^{K} R_{j}$ to reduce variance within a set of $K$ rollouts. 
While effective in many RL tasks, in physics-in-the-loop ILT we observe: 
\begin{itemize}
	\item High reward variance and multi-modality across $\{\vec{q}_{k}\}$ due to non-linear ILT dynamics and mask binarization.
	\item Sensitivity of the group mean to outliers, which destabilizes the baseline.
	\item Weak advantages when most samples are similarly good or bad, leading to slow learning. 
	\item Coupling among rollouts that complicates distributed training. 
\end{itemize}
Instead, we use a frozen teacher $G_{\text{T}}$ (the pretrained model) evaluated on the \emph{same} $\vec{Z}$ and latents $\{\vec{q}_{k}\}$ to obtain a self-critical, teacher-relative baseline: 
\begin{align}
	A_{k} = R_{k} - R^{\text{T}}_{k}, 
\end{align}
where,
\begin{align}
	R^{\text{T}}_{k} = -\operatorname{EPE}\!\big(\vec{M}^{\text{ILT}}_{k,\text{T}}, \vec{Z}\big), \quad \vec{M}^{\text{ILT}}_{k,\text{T}} = G_{\text{T}}(\vec{Z}, \vec{q}_{k}).
\end{align}
This per-sample baseline gives stronger, better-aligned credit assignment, reduces variance without relying on the group mean, and preserves the benefits of group sampling ($K$ rollouts per design). It also anchors progress to a known-good policy, yielding more stable updates early in fine-tuning.

\paragraph{Policy loss.}
We treat pixels as independent Bernoulli variables with probabilities $\text{sigmoid}(\vec{Y}_{k})$. 
Using the standard binary cross-entropy (BCE) as a surrogate for $-\log P$, the policy gradient loss is
\begin{align}
	\mathcal{L}_{\text{pg}} 
	= - \,\mathbb{E}_{\vec{q}_{k}}\!\big[ A_{k} \cdot \big(-\operatorname{BCE}(\vec{Y}_{k}, \vec{M}_{k})\big) \big],
\end{align}
where $\vec{M}_{k}=\mathbb{I}[\vec{Y}_{k}>0.5]$ is detached when computing BCE so that gradients flow only through $\vec{Y}_{k}$.

\paragraph{Imitation toward ILT-refined masks.}
We additionally distill the ILT outcome into $G$ via a smoothed $L_{2}$ loss
\begin{align}
	\mathcal{L}_{\text{imit}} 
	= \mathbb{E}_{\vec{q}_{k}} \big[ \lVert \vec{Y}_{k} - \mathcal{S}(\vec{M}^{\text{ILT}}_{k}) \rVert_{2}^{2} \big],
\end{align}
where $\mathcal{S}(\cdot)$ is a mild low-pass smoothing of the refined mask (e.g., 25$\times$25 stride-1 average pooling) to provide a stable target.

\paragraph{Overall objective and practicalities.}
The total loss is
\begin{align}
	\mathcal{L}_{\text{FT}} = \lambda_{\text{pg}} \, \mathcal{L}_{\text{pg}} + \lambda_{\text{imit}} \, \mathcal{L}_{\text{imit}},
\end{align}
optimized with distributed data parallelism and cosine-annealed learning rates. 
By default we use post-ILT rewards (physics-in-the-loop, a few ILT steps at low resolution), 
though an alternative “initial-mask” reward (no ILT) is supported. 

\subsection{Discussion}

\paragraph{RL for image generators versus RL for LLMs.}
RL finetuning for LLMs operates over a relatively low-dimensional action space (tokens within a limited context window), whereas inverse lithography requires decisions over millions of pixels per clip. This gap---often orders of magnitude (e.g., 128K tokens vs. multi-megapixel images)---explains why RL is rarely used to train image generators. Our style-aware sampler addresses this barrier by moving exploration to a compact latent/style space that preserves design semantics, enabling stable group sampling and physics-grounded updates in a regime that is otherwise intractable.

\paragraph{Reward design is flexible.}
We use post-ILT EPE as a canonical reward because it directly reflects placement fidelity. The same interface naturally supports practical lithography factors, including process variation (PV) bands, depth of focus (DoF), line-edge/width roughness (LER/LWR), mask complexity/shot count, and manufacturability constraints. Multiple factors can be aggregated (e.g., weighted sums) without modifying the learning algorithm, allowing task-specific objectives and easy policy retargeting.

\paragraph{Stability and variance reduction.}
Physics-in-the-loop RL can be noisy and unstable. We mitigate this with (i) strong generative pretraining, (ii) a \emph{teacher-relative} per-sample baseline that anchors progress to a frozen policy on the same latents, and (iii) an imitation term that distills short-ILT refinements into the generator. Together with group sampling, these choices reduce gradient variance, improve sample efficiency, and avoid collapse.

\paragraph{Throughput and cost.}
Embedding ILT in the reward is computationally heavier than purely learned proxies. However, GPU-accelerated implementations and our low-resolution, few-step refinement loop keep wall-clock overhead modest. Batching and distributed data parallelism further amortize cost, making physics-grounded rewards practical during finetuning.

\paragraph{Scalability and deployment.}
The framework accepts arbitrary input sizes and scales to full-chip by tiling or large production tiles. Training and inference both benefit from distributed execution, and the sampler’s design-faithful proposals integrate seamlessly with existing ILT toolchains.

\paragraph{Compute–quality scaling.}
The method exposes clear knobs, e.g. number of proposals per design, ILT steps, and reward fidelity, that trade compute for mask quality. In practice, we observe a favorable scaling trend typical of non-convex, ill-posed problems: additional compute reliably pushes solutions toward better optima, giving users controllable accuracy–throughput trade-offs under a unified framework.

\section{Results}
\label{sec:result}

\subsection{Dataset and Configurations}
To support the claims and our methodologies, we use the popular lithography benchmarks from \lithobench \cite{lithobench} and \texttt{ICCAD13} CAD contest \cite{OPC-ICCAD2013-Banerjee},
where the former contains four subsets termed MetalSet, ViaSet, StdMetal and StdContact.
The statistics are: MetalSet 14{,}824; ViaSet 104{,}773; StdMetal 271; StdContact 165; ICCAD13 10. 
Specifically, we pretrain on MetalSet and ViaSet, and test on StdMetal, StdContact, and ICCAD13.
Since there is no dedicated training set aligned with the ICCAD13 benchmark, the 10 ICCAD13 cases are out-of-distribution (OOD); we show that our method generalizes to these OOD cases by learning numerical ILT behaviors via RL.
Our code is implemented in PyTorch~2.3.0 with CUDA~12.4 and runs on a single DGX node with 8$\times$A100@80GB GPUs using Distributed Data Parallel (NCCL backend).

\paragraph{Pretraining setup.}
We pretrain the generator for 50 epochs using a the objectives in \Cref{eq:pt-objective}.
The discriminator uses Adam (2e{-}4, betas 0.5/0.999), while the generator uses Prodigy with cosine annealing.
Batch size is 16 with 16 data-loading workers under DDP.

\paragraph{RL finetuning setup.}
We initialize from a pretrained inverse model and finetune on the training dataset (image size 2048$\times$2048).
Key hyperparameters: $E{=}20$ epochs, batch size 8 (one design per step), group size $K{=}16$ proposals per design, latent dimension $z{=}256$, learning rate $1{\times}10^{-4}$ (Prodigy), cosine LR schedule (minimum $1{\times}10^{-7}$).
Physics-in-the-loop reward runs a short ILT loop at low resolution (downsample factor 8) for 100 iterations, upsamples via a custom low-resolution pooling followed by bicubic interpolation, and binarizes at 0.5 before evaluation.
We use post-ILT EPE as reward, a self-critical baseline given by the group mean, and an imitation term (LpLoss, $p{=}2$) toward the ILT-refined masks with 25$\times$25 smoothing.
Loss weights: policy gradient $\lambda_{\text{pg}}{=}500$, imitation $\lambda_{\text{imit}}{=}1$.

Evaluation follows prior ILT settings; during training we adopt a stricter EPE violation threshold of 3\,$nm$ to encourage harder finetuning.
we have motivations: (i) 15\,$nm$ is overly permissive even at the 45\,$nm$ node; (ii) targeting 0\,EPE yields vanishing or unstable policy-gradient signals.
We perform morphological opening and cleanup to ensure curvilinear MRC compliance before printability evaluation.

\begin{figure*}[t]
    \centering
    \setlength{\tabcolsep}{3pt}
    \begin{tabular}{ccccccc}
    \begin{minipage}{0.11\textwidth}\centering
    \includegraphics[width=\linewidth]{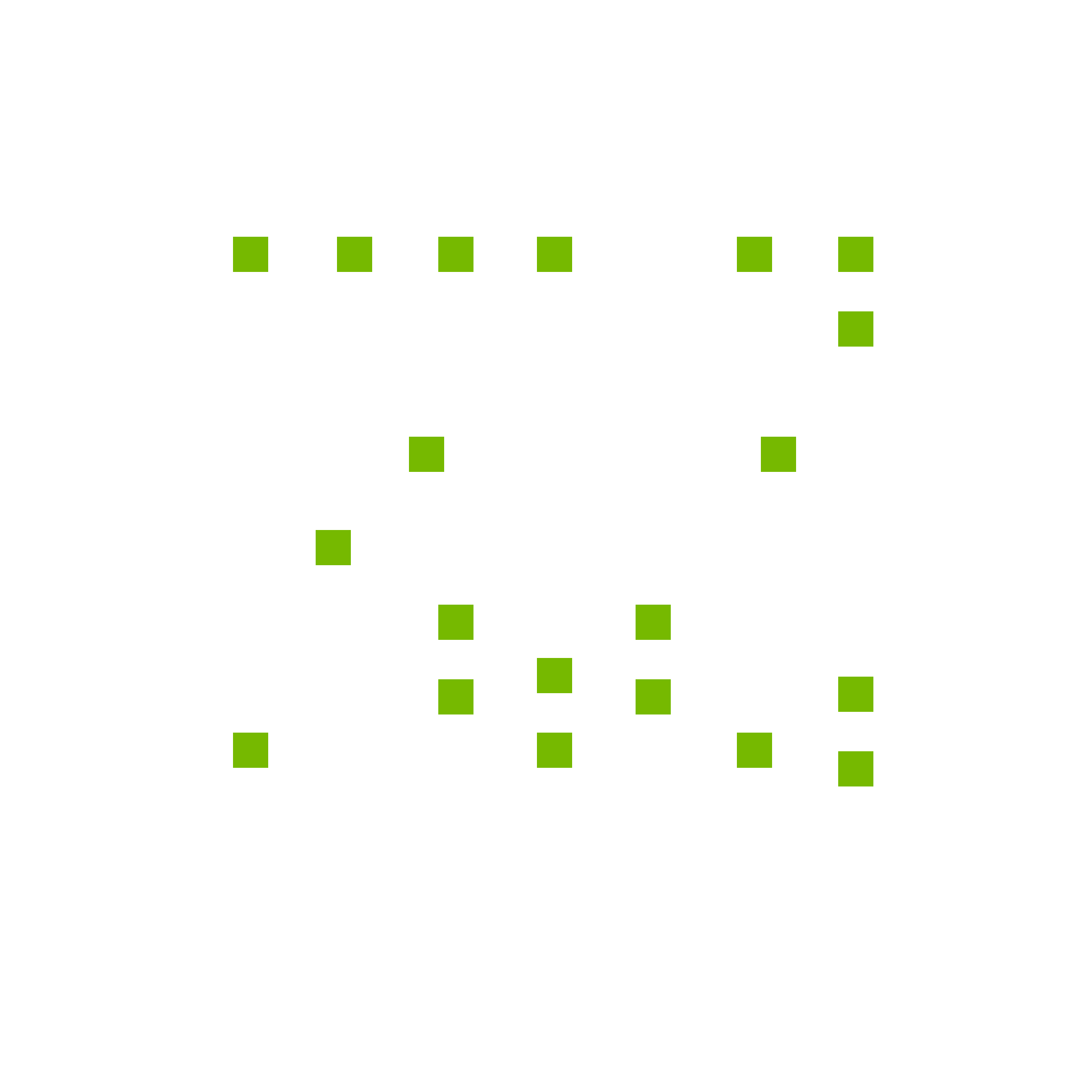}\\
    {\scriptsize Target}
    \end{minipage} &
    \begin{minipage}{0.11\textwidth}\centering
    \includegraphics[width=\linewidth]{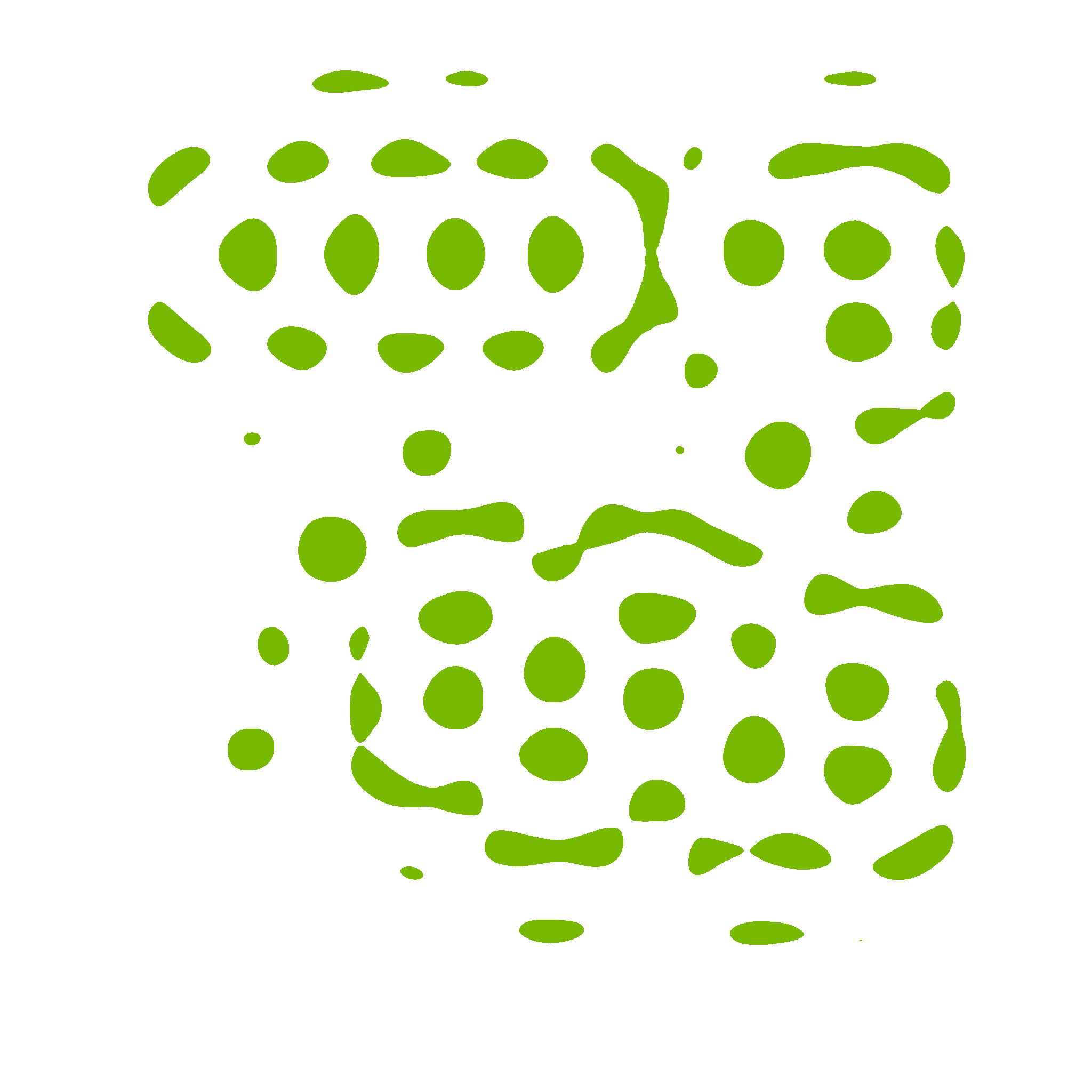}\\
    {\scriptsize CurvyILT Mask}
    \end{minipage} &
    \begin{minipage}{0.11\textwidth}\centering
    \includegraphics[width=\linewidth]{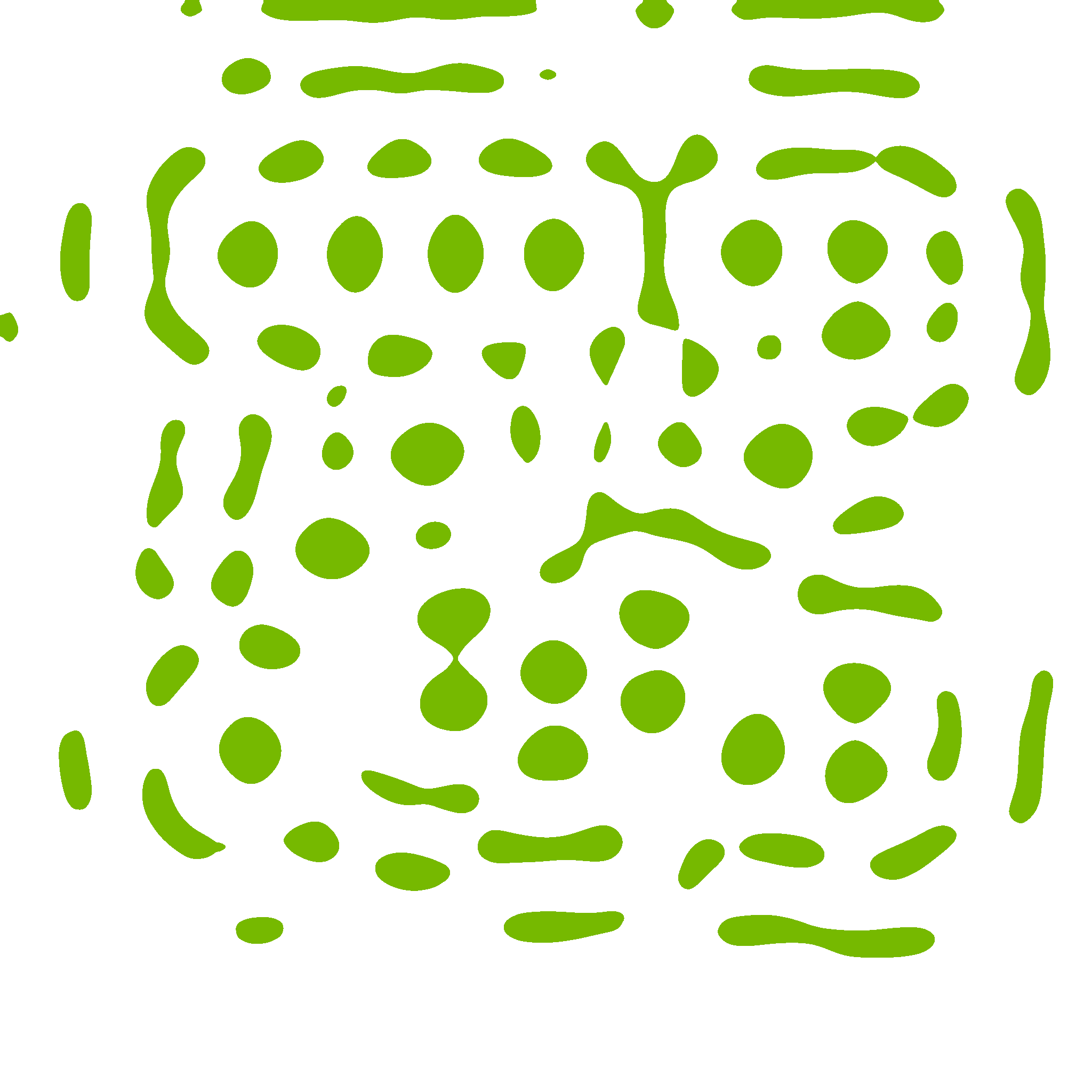}\\
    {\scriptsize Posterior Sample 1}
    \end{minipage} &
    \begin{minipage}{0.11\textwidth}\centering
    \includegraphics[width=\linewidth]{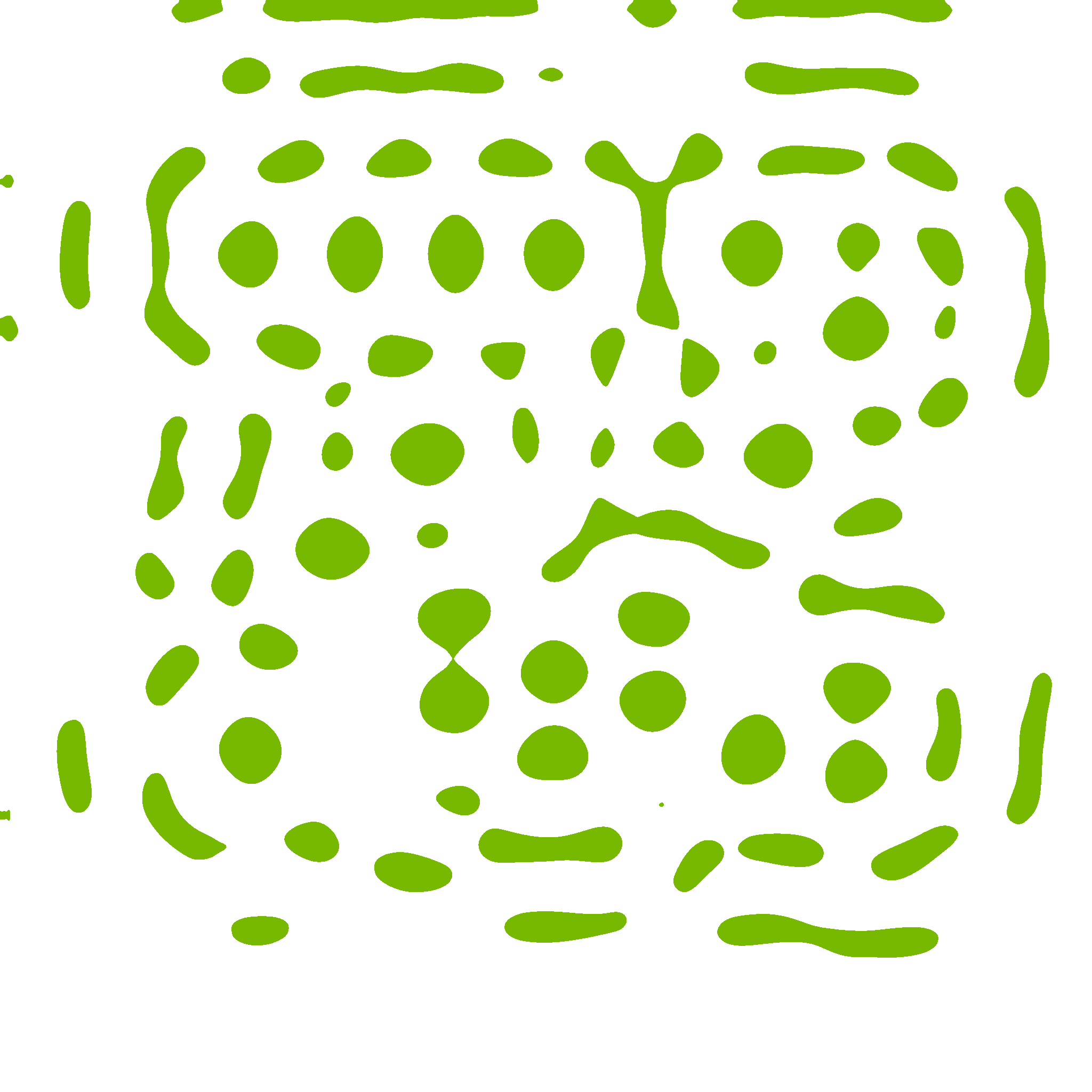}\\
    {\scriptsize Posterior Sample 2}
    \end{minipage} &
    \begin{minipage}{0.11\textwidth}\centering
    \includegraphics[width=\linewidth]{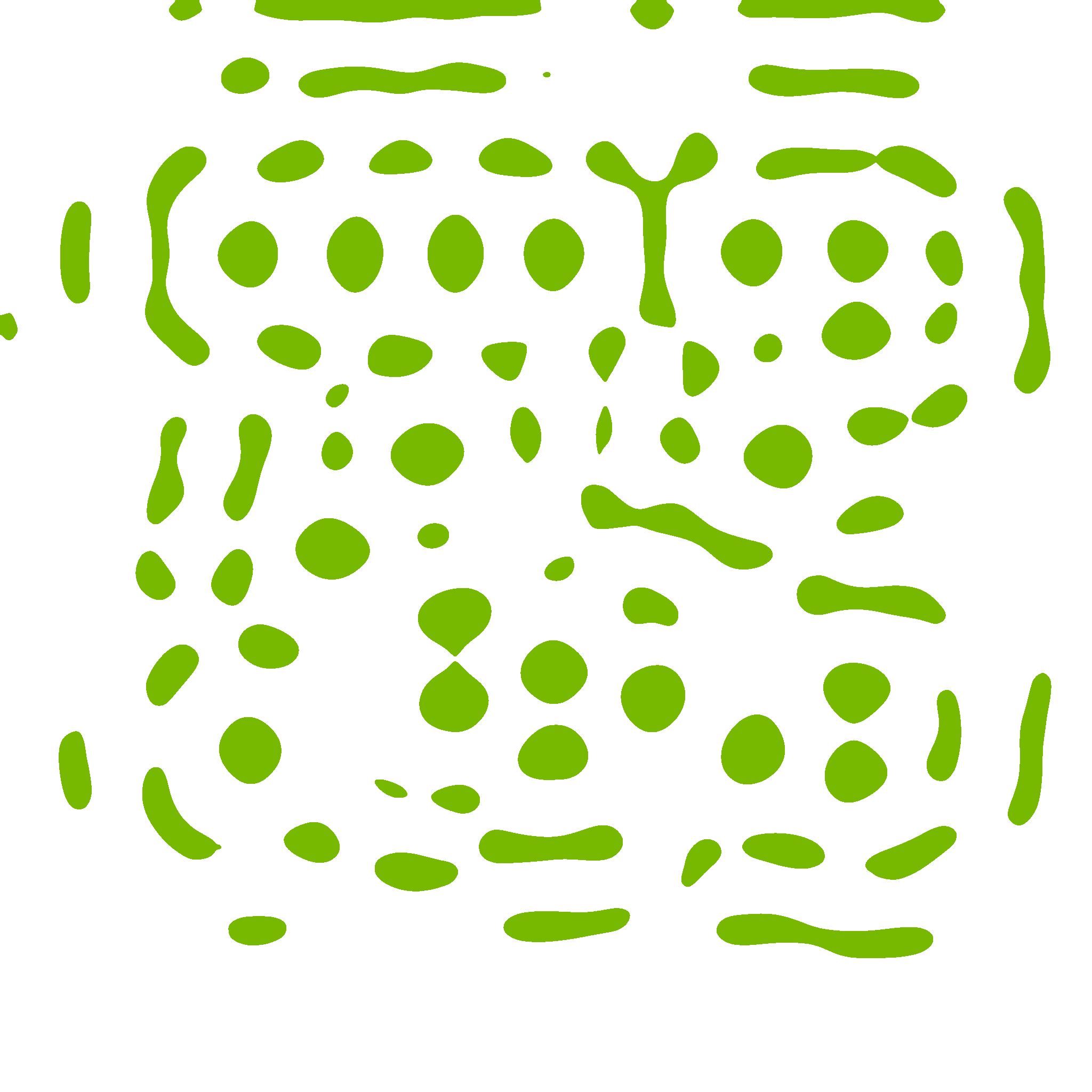}\\
    {\scriptsize Posterior Sample 3}
    \end{minipage} &
    \begin{minipage}{0.11\textwidth}\centering
    \includegraphics[width=\linewidth]{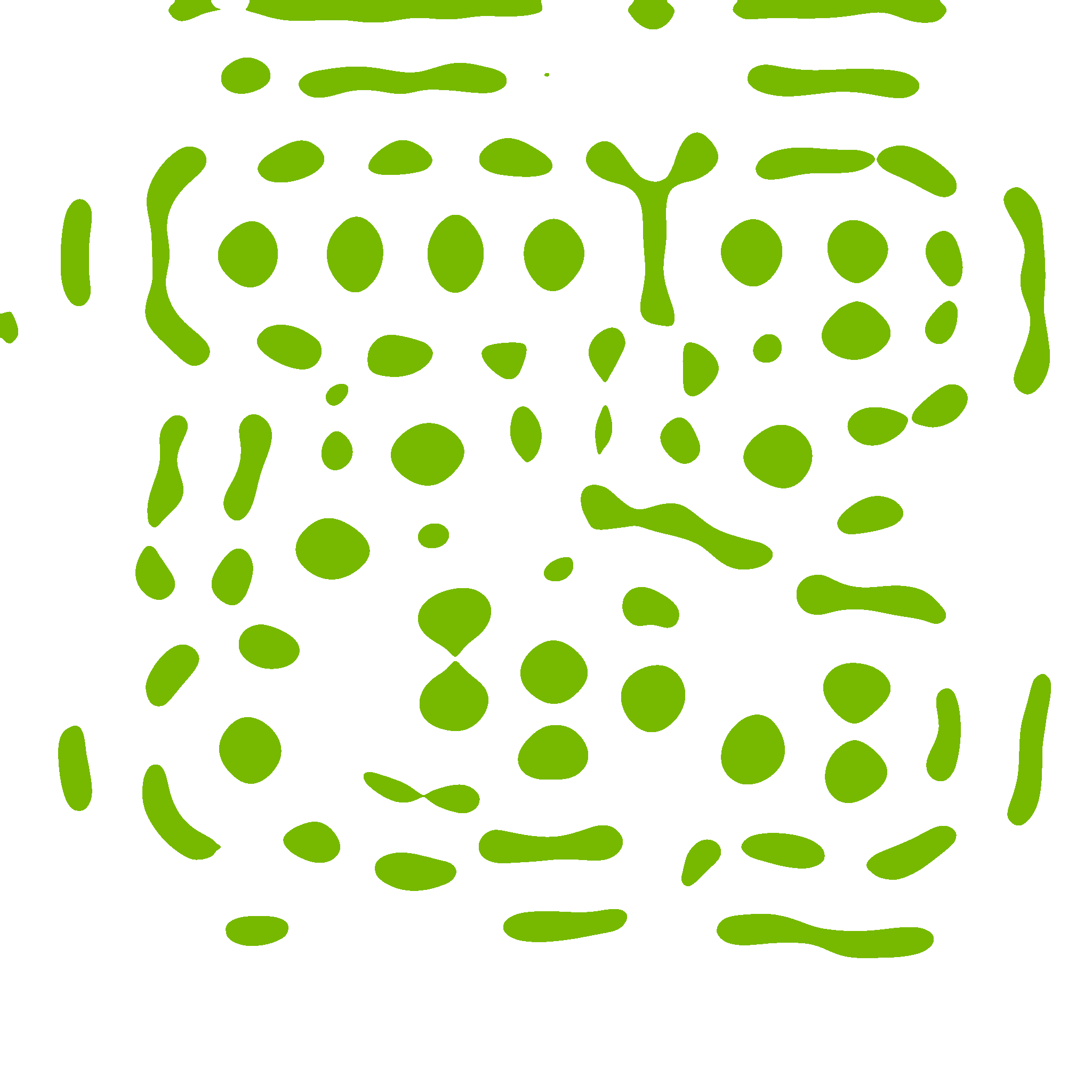}\\
    {\scriptsize Posterior Sample 4}
    \end{minipage} &
    \begin{minipage}{0.11\textwidth}\centering
    \includegraphics[width=\linewidth]{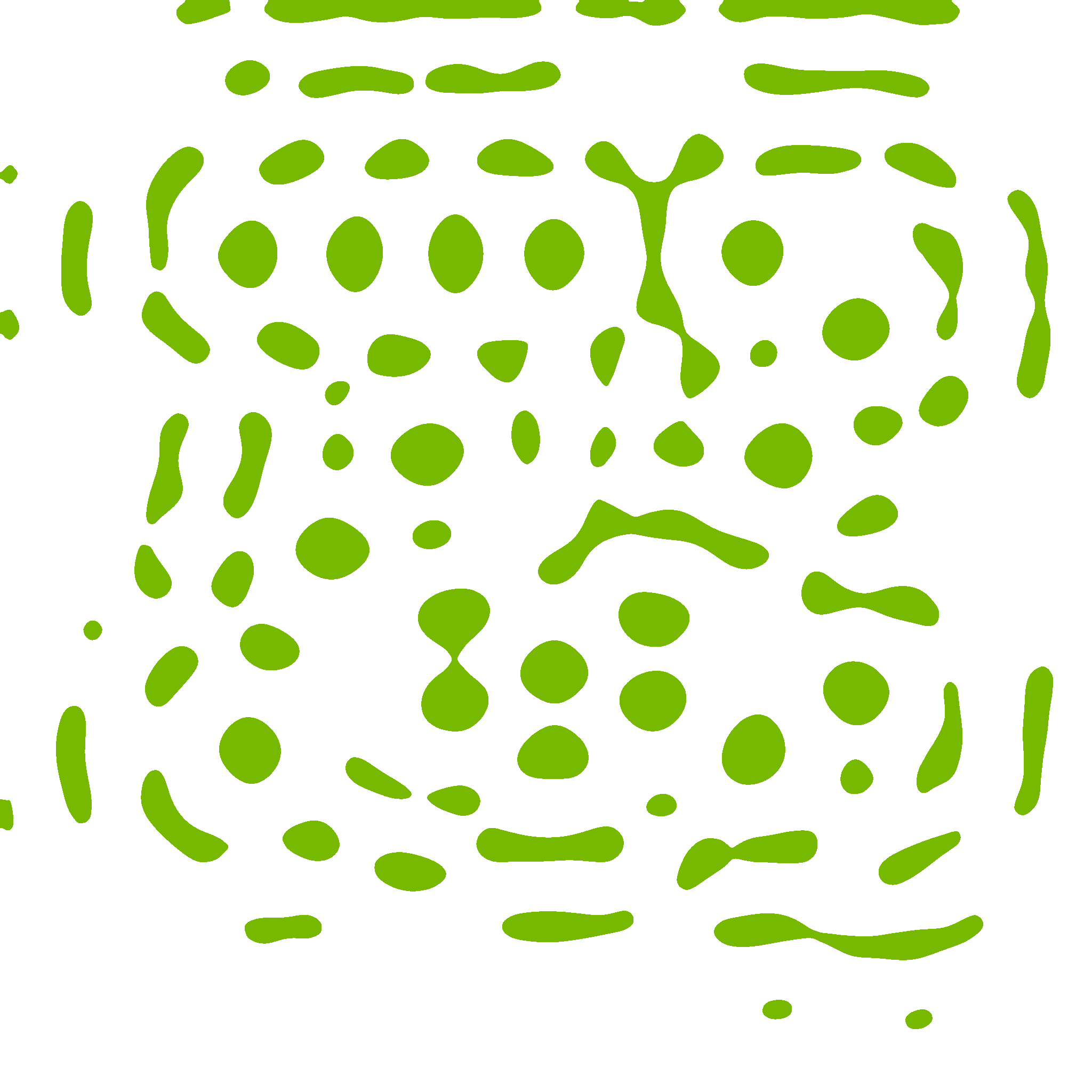}\\
    {\scriptsize Posterior Sample 5}
    \end{minipage}
    \\[0.4em]
    \begin{minipage}{0.11\textwidth}\centering
    \vspace{0pt}
    \end{minipage} &
    \begin{minipage}{0.11\textwidth}\centering
    \includegraphics[width=\linewidth]{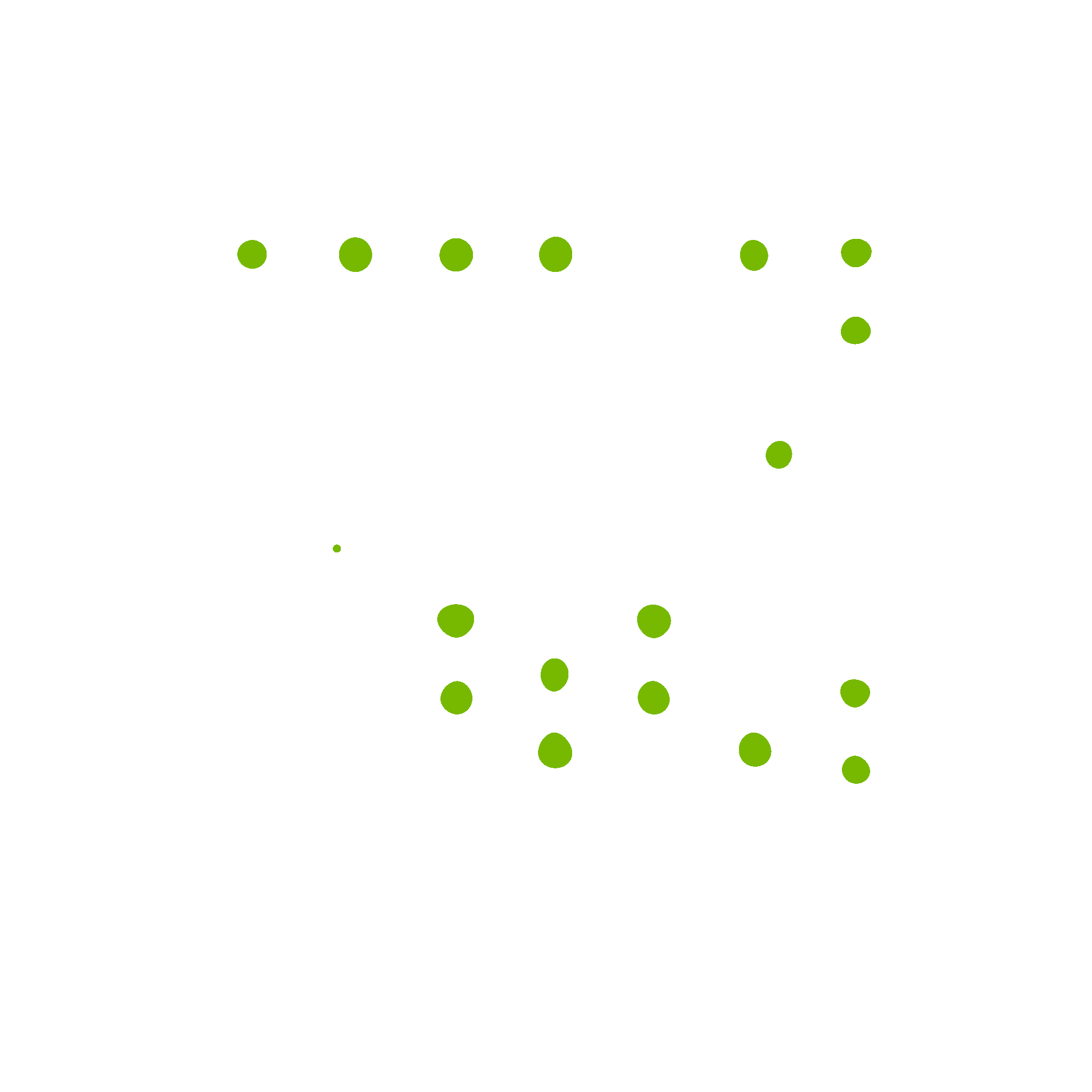}\\
    {\scriptsize EPE=36}
    \end{minipage} &
    \begin{minipage}{0.11\textwidth}\centering
    \includegraphics[width=\linewidth]{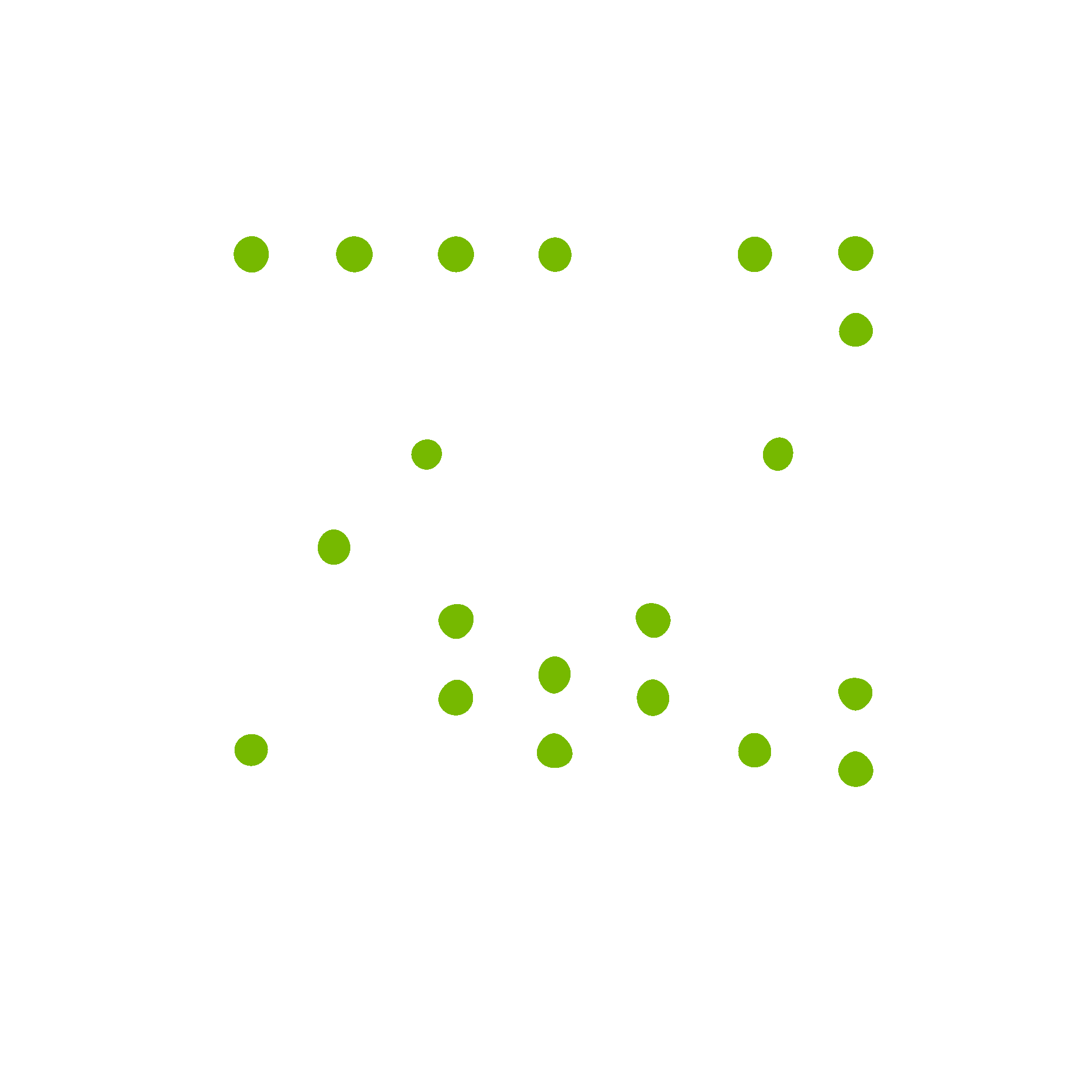}\\
    {\scriptsize EPE=7}
    \end{minipage} &
    \begin{minipage}{0.11\textwidth}\centering
    \includegraphics[width=\linewidth]{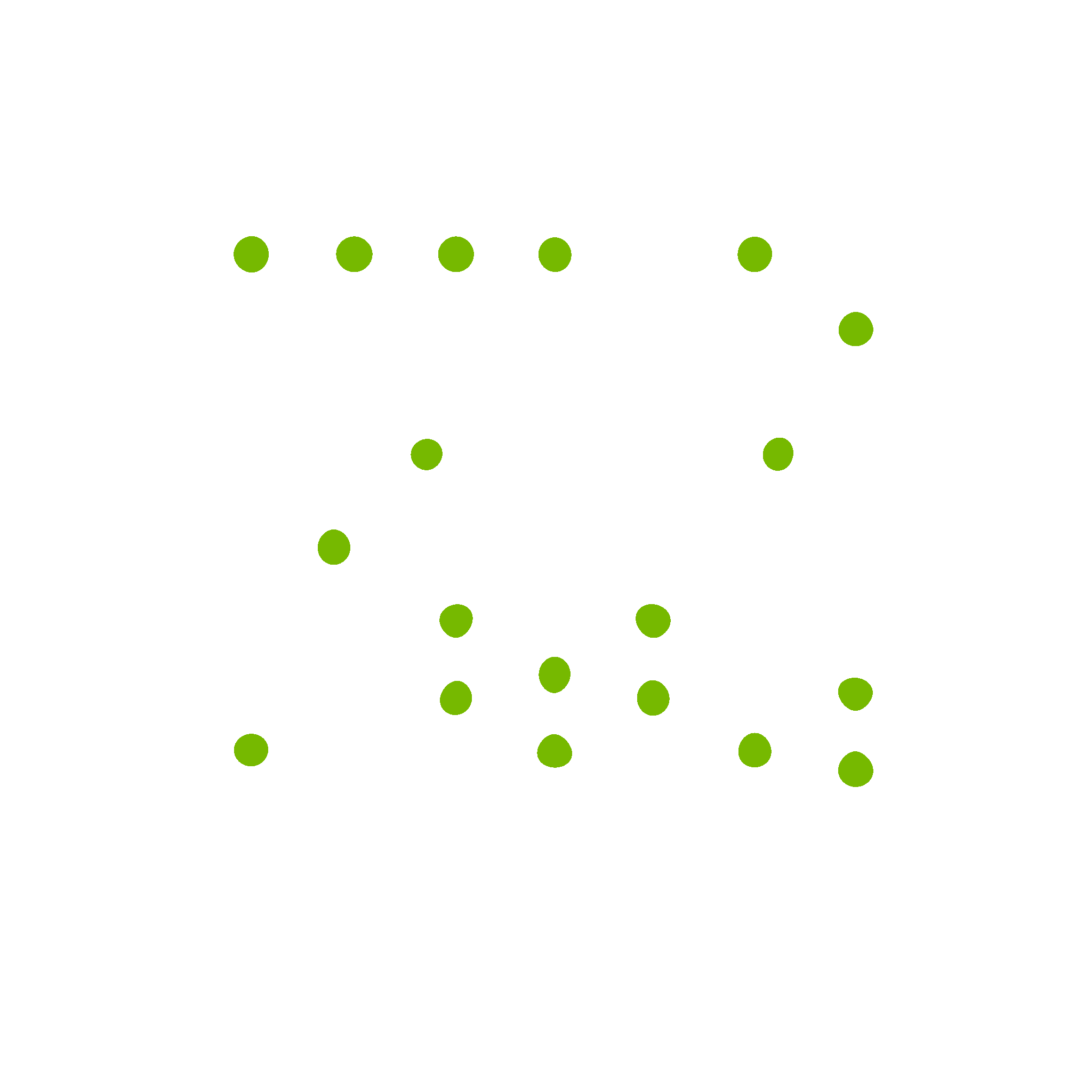}\\
    {\scriptsize EPE=5}
    \end{minipage} &
    \begin{minipage}{0.11\textwidth}\centering
    \includegraphics[width=\linewidth]{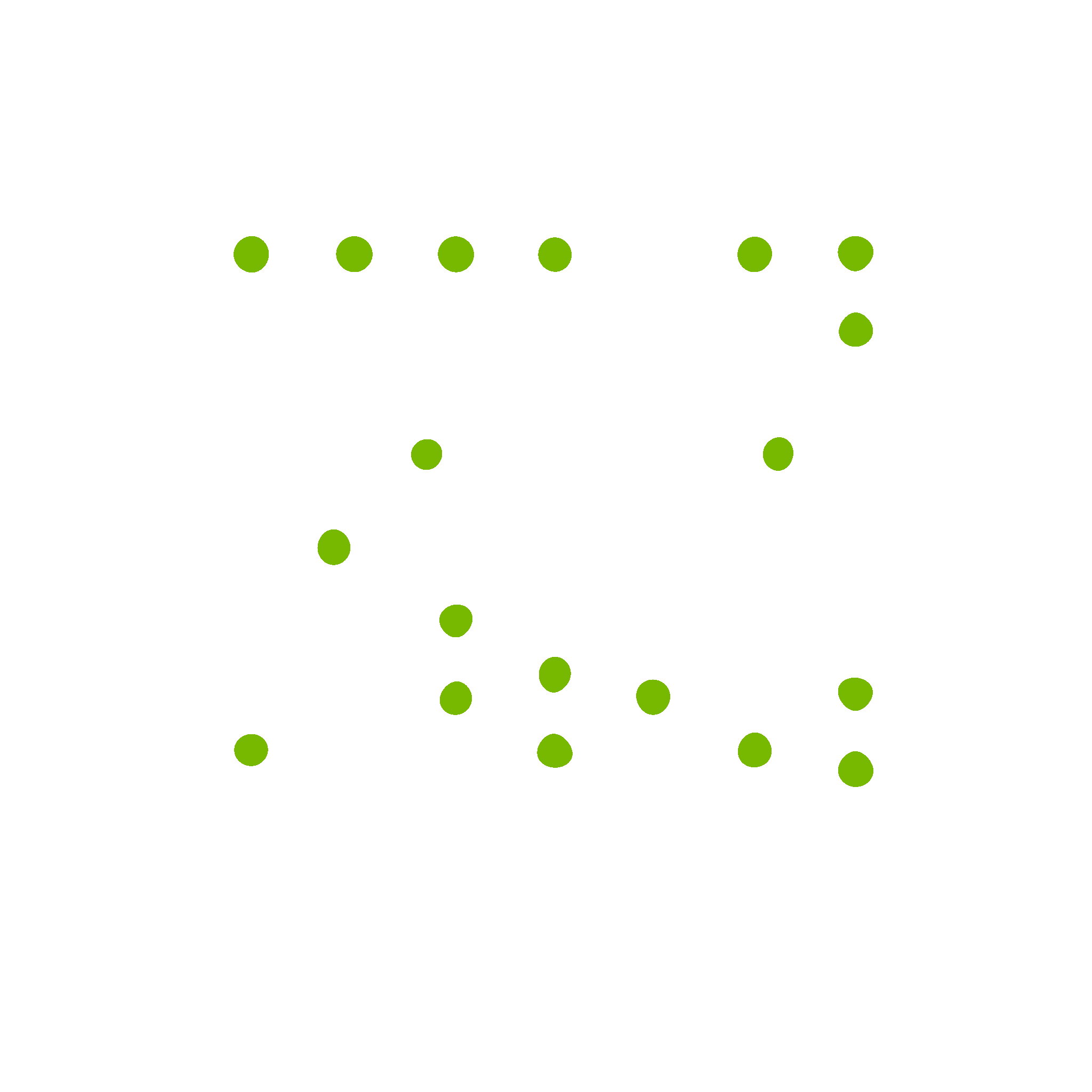}\\
    {\scriptsize EPE=2}
    \end{minipage} &
    \begin{minipage}{0.11\textwidth}\centering
    \includegraphics[width=\linewidth]{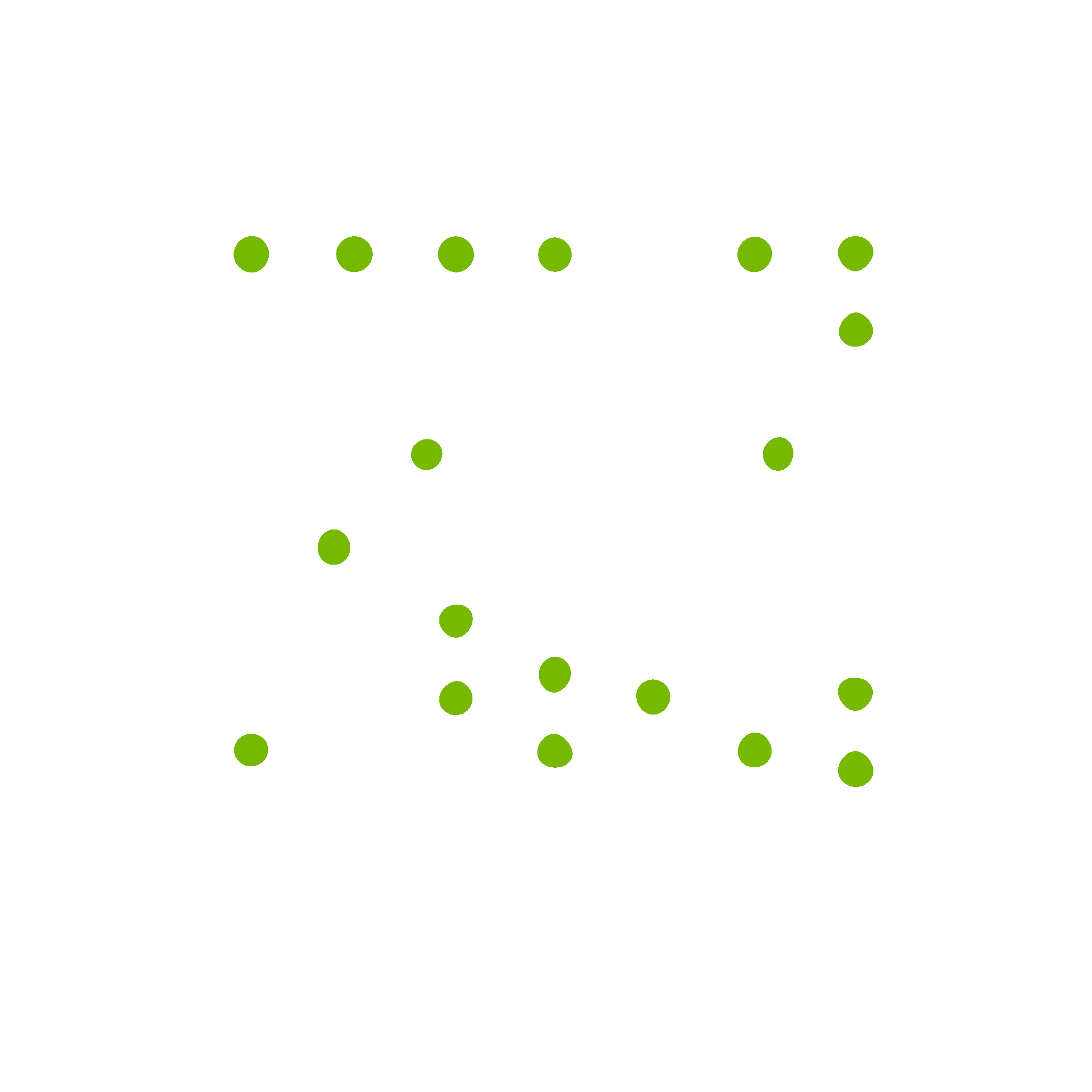}\\
    {\scriptsize EPE=6}
    \end{minipage} &
    \begin{minipage}{0.11\textwidth}\centering
    \includegraphics[width=\linewidth]{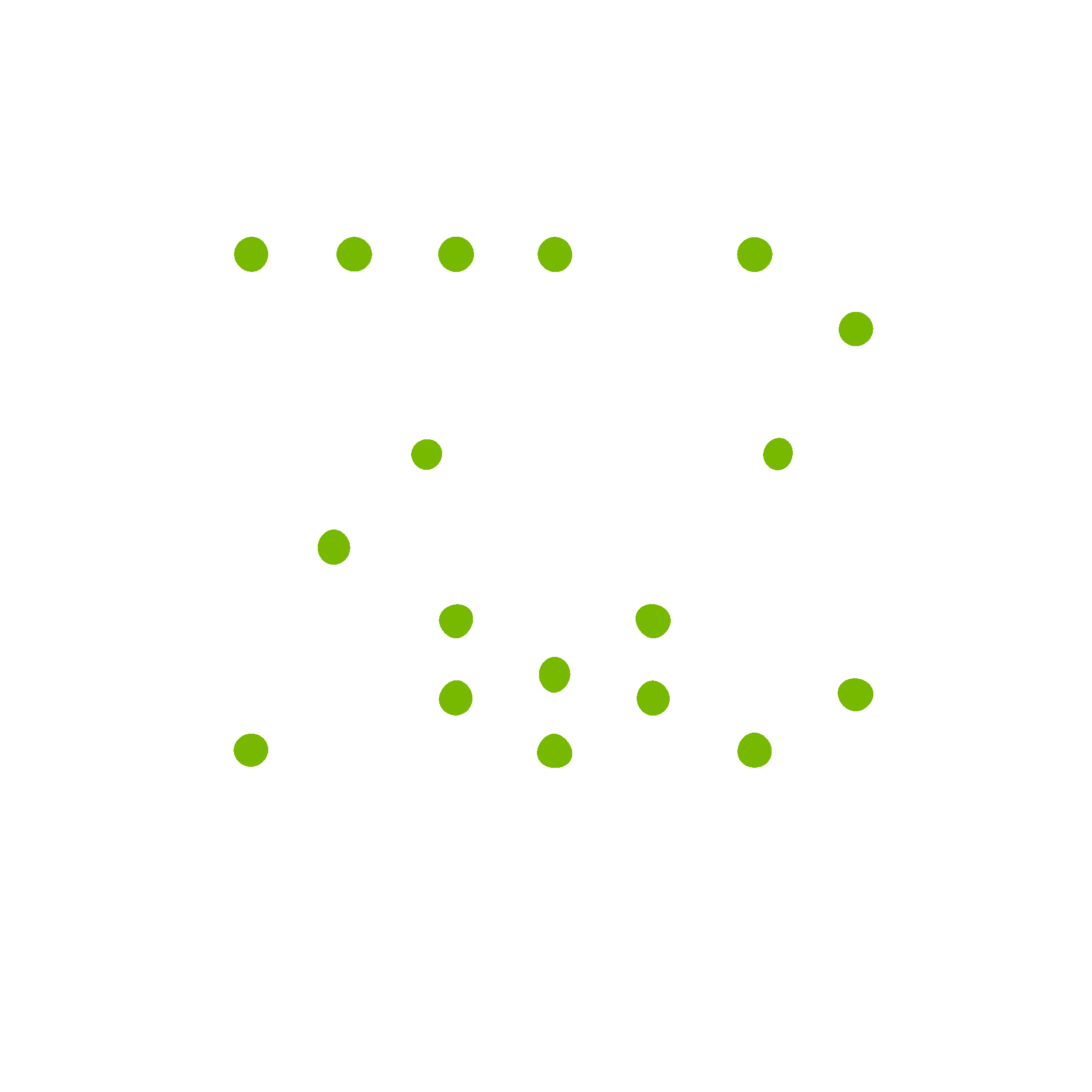}\\
    {\scriptsize EPE=11}
    \end{minipage}
    \end{tabular}
    
    \caption{Visualization of our methods for an example design (HA\_X1\_\_1\_0). Left: target. Top row: golden mask (EPE in nm) and five posterior sampled masks from $G$. Bottom row: corresponding nominal images.}
    \hspace{.1cm}
    \label{fig:examples_ha_x1_1_0}
    \end{figure*}

\subsection{Comparison with State-of-the-Art}
In the first experiment, we compare against recent ILT solvers on standard benchmarks; results are summarized in \Cref{tab:result15}. Our method achieves consistently lower or matching EPE while remaining competitive on PV. On StdContact (Avg), we reduce EPE from 8.6 (DAC'23) and 3.8 (ISPD'25) to 2.0 with comparable PV. On StdMetal (Avg), all methods reach 0.0 EPE, and we obtain the best PV (21029.2). On ICCAD13, our average EPE is the lowest (1.6 vs.\ 1.8 for ISPD'25 and 2.5 for DAC'23), with PV close to ISPD'25 and better than DAC'23. Notably, we achieve these gains using roughly half of the iteration budget (and thus runtime) by limiting the short, low-resolution ILT refinement inside the loop, highlighting the efficiency of reinforcement finetuning with a style-aware sampler.

\begin{table*}[t]
\centering
\caption{Result comparison with state-of-the-art ILT solvers under 15$nm$ EPE threshold.}
\label{tab:result15}
 \setlength{\tabcolsep}{10pt}
 \begin{tabular}{lrrrrrrrr}
\toprule
 & \multicolumn{2}{c}{DAC'22 \cite{OPC-DAC2022-Wang}} & \multicolumn{2}{c}{DAC'23 \cite{OPC-DAC2023-Sun}} & \multicolumn{2}{c}{ISPD'25-300 iterations \cite{curvyilt}} & \multicolumn{2}{c}{Ours-150 iterations} \\
\cmidrule(lr){2-3}\cmidrule(lr){4-5}\cmidrule(lr){6-7}\cmidrule(lr){8-9}
Benchmarks & EPE & PV & EPE & PV & EPE & PV & EPE & PV \\
\midrule
StdContact-Avg & - & - & 8.6 & 39997.0 & 3.8 & 36172.0 & \textbf{2.0} & 39186.4 \\
StdMetal-Avg & - & - & 0.0 & 24928.0 & 0.0 & 21631.0 & 0.0 & 21029.2 \\
ICCAD13-1 & 7 & 47015 & 3 & 47015 & 3 & 44447 & 3 & 47459 \\
ICCAD13-2 & 3 & 37555 & 0 & 37555 & 0 & 36914 & 0 & 33965 \\
ICCAD13-3 & 62 & 69361 & 22 & 69361 & 15 & 70580 & \textbf{13 }& 74370 \\
ICCAD13-4 & 2 & 21514 & 0 & 21514 & 0 & 21584 & 0 & 21985 \\
ICCAD13-5 & 1 & 49683 & 0 & 49683 & 0 & 47870 & 0 & 47781 \\
ICCAD13-6 & 2 & 44127 & 0 & 44127 & 0 & 42288 & 0 & 42987 \\
ICCAD13-7 & 0 & 36961 & 0 & 36961 & 0 & 34389 & 0 & 36062 \\
ICCAD13-8 & 0 & 20985 & 0 & 20985 & 0 & 18649 & 0 & 18312 \\
ICCAD13-9 & 2 & 54948 & 0 & 54948 & 0 & 54387 & 0 & 52970 \\
ICCAD13-10 & 0 & 16581 & 0 & 16581 & 0 & 15014 & 0 & 14916 \\
ICCAD13-Avg & 7.9 & 39873 & 2.5 & 39873 & 1.8 & 38612.2 & \textbf{1.6} & 39080.7 \\
\bottomrule
\end{tabular}
\end{table*}

\subsection{Stress Test on ILT Solver}
To probe robustness under tight manufacturing tolerances, we stress test all solvers at a stringent 3\,nm EPE threshold; results are summarized in \Cref{tab:result3}. 
Relative to CurvyILT (ISPD'25), our pretrained model already cuts EPE substantially while keeping PV comparable (e.g., StdContact-Avg 40.4$\rightarrow$9.0, StdMetal-Avg 11.4$\rightarrow$7.2, ICCAD13-Avg 32.8$\rightarrow$27.3). 
With RL finetuning, EPE improves further across aggregates (StdContact-Avg 6.5, StdMetal-Avg 6.7, ICCAD13-Avg 25.2), outperforming the baseline on the popular benchmarks at this stricter criterion. 
These gains are achieved under identical physics using roughly half the iteration budget (and thus runtime), thanks to our short low-resolution ILT loop guided by the learned sampler.
The superior performance of the posterior sampled masks also demonstrate the effectiveness of our methodology, offering (almost) free scaling of optimization performance.

\begin{table}[t]
\centering
\caption{Stress-test at 3$nm$ EPE threshold comparing CurvyILT (ISPD'25, 300 iterations) and our methods (150 iterations).}
\label{tab:result3}
\setlength{\tabcolsep}{4pt}
\begin{tabular}{lrrrrrr}
\toprule
 & \multicolumn{2}{c}{ISPD'25 \cite{curvyilt}} & \multicolumn{2}{c}{Ours (PT)} & \multicolumn{2}{c}{Ours (PT+RL)} \\
\cmidrule(lr){2-3}\cmidrule(lr){4-5}\cmidrule(lr){6-7}
Benchmarks & EPE & PV & EPE & PV & EPE & PV \\
\midrule
StdContact-Avg & 40.4 & 32969.6 & 9.0 & 39160.4 & \textbf{6.5} & 39907.5 \\
StdMetal-Avg & 11.4 & 19083.8 & 7.2 & 21507.6 & \textbf{6.7}& 21029.2 \\
ICCAD13-1 & 51 & 44446 & 47 & 48098 & \textbf{47} & 47459 \\
ICCAD13-2 & 34 & 36940 & 39 & 40330 & \textbf{29} & 36948 \\
ICCAD13-3 & 107 & 70545 & 97 & 68814 & \textbf{86} & 74370 \\
ICCAD13-4 & 8 & 21577 & 7 & 24624 & \textbf{6} & 22647 \\
ICCAD13-5 & 29 & 47861 & 15 & 51175 & 16 & 50921 \\
ICCAD13-6 & 28 & 42287 & 21 & 44333 & \textbf{21} & 43866 \\
ICCAD13-7 & 7 & 34409 & 3 & 35633 & \textbf{1} & 37091 \\
ICCAD13-8 & 13 & 18644 & 10 & 19283 & \textbf{8} & 19496 \\
ICCAD13-9 & 49 & 54393 & 34 & 56981 & 38 & 55105 \\
ICCAD13-10 & 2 & 15013 & 0 & 15970 & \textbf{0} & 16256 \\
ICCAD13-Avg & 32.8 & 38611.5 & 27.3 & 40524.1 & \textbf{25.2} & 40415.9 \\
\bottomrule
\end{tabular}
\end{table}

\subsection{Result Visualization}
Finally, we visualize posterior samples for the same design, along with their nominal images after ILT refinement.
This illustrates how different initializations lead to different final mask quality under the same optimization budget.
The visualization and prior runs indicate that (i) initialization matters, (ii) RL enables masks unreachable by a purely numerical solver, and (iii) a good initial mask does not necessarily yield the best refined mask.

\section{Conclusion}
\label{sec:conclu}
We presented a physics-grounded RL framework for inverse lithography that treats the generator as a design‑conditioned sampler and couples it with short, low‑resolution ILT in the loop. A style‑aware architecture and a policy+imitation objective let physics define the reward while distilling ILT outcomes back into the model. On standard benchmarks we achieve state‑of‑the‑art EPE at 15\,nm and remain robust at 3\,nm, often with about half the iteration budget and competitive PV. The approach scales to arbitrary sizes, integrates with existing ILT toolchains, and naturally extends beyond EPE (PV, DoF, LER/LWR, mask complexity). Future work will pursue richer multi‑objective rewards and faster physics to further improve accuracy and throughput.

{
    \bibliographystyle{IEEEtran}
    \bibliography{ref/Top,ref/DFM,ref/Additional,ref/LLM,ref/PD,ref/HSD}
}

\end{document}